# Grad-Instructor: Universal Backpropagation with Explainable Evaluation Neural Networks for Meta-learning and AutoML

Ryohei Ino

## Abstract

This paper presents a novel method for autonomously enhancing deep neural network training. My approach employs an *Evaluation Neural Network (ENN)* trained via deep reinforcement learning to predict the performance of the target network. The ENN then works as an additional evaluation function during backpropagation.

Computational experiments with Multi-Layer Perceptrons (MLPs) demonstrate the method's effectiveness. By processing input data at $0.15^2$ times its original resolution, the ENNs facilitated efficient inference. Results indicate that MLPs trained with the proposed method achieved a mean test accuracy of 93.02%, which is 2.8% higher than those trained solely with conventional backpropagation or with L1 regularization. The proposed method's test accuracy is comparable to networks initialized with He initialization while reducing the difference between test and training errors. These improvements are achieved without increasing the number of epochs, thus avoiding the risk of overfitting.

Additionally, the proposed method dynamically adjusts gradient magnitudes according to the training stage. The optimal ENN for enhancing MLPs can be predicted, reducing the time spent exploring optimal training methodologies.

The explainability of ENNs is also analyzed using Grad-CAM, demonstrating their ability to visualize evaluation bases and supporting the Strong Lottery Ticket hypothesis.

# 1. Introduction

 Backpropagation (BP) achieves unprecedented learning speed in the training of deep neural networks (DNNs), contributing to the generalizability and accuracy of DNNs. Methods to manipulate the gradient to prevent overlearning, improve accuracy, and accelerate learning include L1 regularization (Anders Krogh & John A. Hertz), batch normalization (Sergey Ioffe & Christian Szegedy, 2015), warmup (Prin Phunyaphibarn et al, 2023), SGDR (Ilya Loshchilov & Frank Hutter, 2016), Adam (Kingma & Ba, 2014), Lion (Xiangning Chen et al., 2023), and many other methods have been proposed. However, the manual search of the methods and hyperparameters to be used requires an enormous amount of effort. Therefore, an approach called Automated Machine Learning (AutoML), which automatically searches for DNN structures and hyperparameters, has been investigated (Xin He et al., 2021). However, AutoML requires training multiple DNNs to search a large number of combinations, which consumes a large amount of computational resources.

 In addition, when training a DNN for a new task, the lack of train data and the time required for training are often problems. Without sufficient training data, it is difficult for DNNs to optimize without overfitting. In such cases, meta-learning (or learning to learn), which uses knowledge from similar datasets or tasks to optimize, has been investigated (Timothy Hospedales et al., 2022). In this paper, DNNs will be referred to as NNs.

 In this study, I propose a novel method for AutoML and meta-learning that allows NNs to learn and improve NNs. The proposed method uses deep reinforcement learning to train an Evaluation Neural Network (ENN) that predicts the performance of the target based on arbitrary parameters representing the characteristics of the target to be improved, and performs BP using this ENN as the evaluation function. Therefore, the proposed method is fast and can optimize parameters regardless of whether they are differentiable or not. The proposed method can theoretically be used in conjunction with existing explanatory methods to visualize the basis of the evaluation neural network, and can also be processed in parallel with conventional BP.

 In this study, I verify through computational experiments that the ENN of the proposed method can learn the relationship between the behavior and performance of the NN to be improved, and can autonomously acquire the ability to improve it. In the experiments, I use a convolutional neural network (CNN) (Saad Albawi et al., 2017) with a convolutional layer and a fully connected layer as the ENN, and a multilayer perceptron (MLP) (Marius-Constantin Popescu et al., 2009) using stochastic gradient descent (SGD) (Sebastian Ruder, 2016) as the optimizer. I also experiment with ENN and Grad-CAM (Ramprasaath R. Selvaraju et al., 2017) together to test if it is possible to visualize the basis of ENN decisions and evaluations. Experimental results show that, in some cases, the post-training test accuracy with the proposed method is 2.8% higher than that of MLPs trained only with regular SGD.

## 2. Universal Backpropagation & Evaluation Neural Networks

In this chapter, the advantages and features of the proposed method are described, including its differences from existing reinforcement learning methods. Next, the concept of the proposed method as an optimization technique is explained with examples of its implementation.

### 2.1. Universal Backpropagation as an optimization method

The author conceived the proposed method as an approach to improve issues related to computational speed and memory consumption in AutoML using meta-learning method.

In the proposed method, some kind of trained NN is used as an evaluation function to conduct BP on the NNs to be improved. I call the trainable evaluation function as Evaluation Neural Network (ENN) in this paper. The parameters of NN to be improved are input to ENN and ENN output predicted value of some kinds of indices, such as test accuracy. I aim to significantly reduce the number of executions of training networks and memory needed by utilizing the exceptional combination search capability of BP.

Using the proposed method with ENN that not overfitted, even for Hyperparameters where BP would normally be impossible, BP can be made feasible. However, when executing BP with the proposed method, optimizing the learning rate for gradient from ENN cannot be automated solely by the proposed method.

Equations (1) and (2) are an example implementation of the proposed method, where P represents any parameter to be improved, W represents parameters can be optimized with conventional BP method ($P \ni W$), and $lr_1$ and $lr_2$ represent learning rate for each. In this implementation example, ENN outputs the predicted rewards of the improvement target. Therefore, gradient ascent method should be used to optimize P.

$$\Delta W = -lr_1 \nabla L \ \ ...(1)$$
$$\Delta P = lr_2 \nabla ENN(P) \ \ ...(2)$$

In this case, parallel processing is possible with both BP using the proposed method and conventional BP. This is because ENN used the proposed method does not require information of NNs' training loss and gradient, thus enhancing processing speed. Moreover, theoretically, performance degradation due to overfitting should be probabilistically mitigated during the BP using the proposed method as far as the ENN outputs the estimate values of test accuracies.

Furthermore, by combining the proposed method with existing visualizing method like grad-CAM, the ENN can explain the contribution of each area to the predicted indices. This can be used to analyze the NN to be improved, allowing identify malfunctioning areas while conducting BP method.

## 2.2. Universal Backpropagation as an AutoML method

Several methods based on Neural Architectural Search (NAS) have been proposed as AutoML methods (Thomas Elsken et al., 2019); however, the proposed method can be more memory-efficient during training compared to NAS. For example, Transformers (Vaswani et al., 2017), unlike CNNs, are not provided with much information on image recognition tasks by humans. In other words, (Vision) Transformers have lower inductive bias than CNNs (Alexey Dosovitskiy et al., 2021). Therefore, they require more training data than CNNs but offer high versatility, such as their applicability in natural language processing. In the proposed method, by preparing one target with low inductive bias, like a Transformer and one ENN, AutoML can be executed through the ENN designing the target. Consequently, this method is likely to be more memory-efficient than NAS, which involves selecting the optimal combination from numerous layers or blocks.

Additionally, a method called AutoGrow (Wei Wen et al., 2020), which optimizes the number of layers, has been proposed as an AutoML technique. With the proposed method, optimization of number of layers can be achieved by allowing the ENN to design the inclusion or exclusion of skip connections. Unlike AutoGrow, the proposed method also incorporates aspects of meta-learning, enabling AutoML to be executed through statistically learned approaches.

## 2.3. Universal Backpropagation as a reinforcement learning method

Deep reinforcement learning methods such as Deep Q-Network (DQN) (Volodymyr Mnih et al., 2013) have been proposed. These are off-policy methods that output discrete values. On the other hand, reinforcement learning methods that can probabilistically output a set of continuous values include the Actor-Critical (AC) method (Vijay R. Konda, 2002), which uses the policy gradient method.

The latter trains both the Actor and the Critic, so each needs to learn the strategy and values of actions. In addition, since the Actor learns policy while selecting actions, it is difficult to learn with data accumulated in the past, and data utilization is less efficient than with Q-learning. Some Actor-Critic based architectures, such as Soft Actor-Critic (Tuomas Haarnoja et al., 2018), can be trained off-policy, but since both Actor and Critic are trained, the number of parameters to be optimized is expected to be large. Therefore, it is expected that applying a method like AC, which includes learning of policy, to meta-learning and simultaneously optimizing parameters in the order of 1000 to 10000 would place a significant burden on computational resources.

The proposed method is off-policy method, meaning that the ENN only needs to learn to predict the reward from a given state. The action selection (output of a set of continuous values) is done by BP of the ENN. In other words, since there is no need to learn policy, the size of the output layer of the ENN is also proportional to the number of types of rewards to be predicted (e.g., accuracy and speed of the NN to be improved). Since a large number of parameters tends to increase the VC dimension (Junyi Guan & Huan Xiong, 2023), when trained with the same number of parameters as AC, the proposed method should be more expressive and can predict the appropriate reward in many cases.

Furthermore, the proposed method is expected to be able to guarantee the validity of the output by the BP using ENN as long as ENN predicts rewards accurately. This is because the function that outputs actions in the proposed method is the derivative of the function that predicts rewards. For example, if the ENN of the proposed method is not able to correctly predict the reward obtained from a certain state, the output by BP is also not valid. Therefore, if a problem such as concept drift occurs in the operation of the proposed method, it is sufficient to implement the continuous learning method (Liyuan Wang et al., 2024) to just correctly predict the reward.

Moreover, since the proposed method also has an aspect of an evaluation function, it is possible to freely change which of the performance indicators to focus on for each improvement goal. This makes the proposed method more versatile than conventional DQN. For example, when using an ENN that predicts the test accuracy a and memory consumption m of the NN to be improved, the final evaluation V can be calculated using equation (3), and this value can be used in the BP calculation to improve the test accuracy rather than reduce the memory consumption.

$$V = 1.3a - 0.3m \ldots (3)$$

## 3. Experiments

ENNs are trained to predict test accuracies from the targets' weight parameters. The ENNs could significantly improve MLPs' accuracies without consuming additional train data while avoiding the risk of overfitting. The ENNs are also able to visualize the basis of their decisions, identifying the MLPs' weights that contribute to the MLPs' accuracies.

### 3.1. Experimental Setup

To investigate whether the proposed method can autonomously learn and improve the training of target MLPs, the MLPs' weight parameters were optimized using the proposed method and conventional BP. The performance metrics measured in this experiment were the test accuracy and the difference between the test error and training error of the target systems. The latter assesses the tendency towards overfitting.

Equations (4) and (5) describe the update rules for weight parameters W used in the experiment. Here, $\bar{y}$ represents the average output of the ENN, $b_t$ is the number of batches trained in that epoch, and $lr_1$ and $lr_2$ denote the learning rates. The $lr_1$ was set to 0.01. In the calculation of $lr_2$, $lr_1$ is divided by $\bar{y}$ to stabilize the gradient magnitude. The ENN used in this experiment predicts the test accuracy of the MLP from the weight parameter array of the MLP to improve it every 10 batches of training the MLP.

In the experiment, each of the four ENNs (ENN S1, ENN S2, ENN B1, ENN B2) optimized six MLPs, and the average performance of these MLPs was measured. For comparison, the performance of training a Vanilla MLP, applying only L1 Regularization, and applying only He initialization were also measured.

During ENNs' training, ENN S1 and B1 distributed the weight parameters of 580 NNs to training and test data, while ENN S2 and B2 distributed those of 694 NNs. The training data for the ENNs included not only dropout, He initialization, and L1 Regularization but also fully connected layers trained by ENNs not used in this experiment. The batch size was 256.

Each ENN consisted of two convolutional layers and two fully connected layers. Both convolutional layers had a filter size of 5×5, with input data channels of 2 and 5, respectively. In the case of ENN S1 and S2, the fully connected layer sizes were 4950×1250 and 1250×1, respectively. Similarly, for ENN B1 and B2, the fully connected layer sizes were 13780×3000 and 3000×1, respectively.

The ENNs' initial values were determined based on a Gaussian distribution with a standard deviation of 0.01. A dropout layer with a dropout rate of 0.2 was placed between the two fully connected layers. The ENNs were trained for 30 epochs using L1 regularization.

Figure 3.1.1 shows the data shape input to the ENN, which is in the form of image data with two channels. In this experiment, the resolution of the input data was reduced: to $0.1^2$ of the original size for ENN S1 and S2, and to $0.15^2$ for B1 and B2. This reduction is necessary because the number of parameters in the NN can exceed the size of the train data. If the parameters of the target NN were directly input into the ENN, the ENN would have many more parameters than the target NN, necessitating a reduction in input resolution. The input data shapes during training were (256×2×78×43) and (256×2×118×64), and during inference were (5×2×78×43) and (5×2×118×64).

The test accuracy of the ENNs was calculated using cosine similarity between predicted and actual values, while test loss and train loss were computed using mean squared error.

The target MLP consisted of five fully connected layers of sizes (784×256), (256×256), (256×256), (256×256), and (256×10), with ReLU functions immediately following each layer. The initial values of the target were determined based on a Gaussian distribution with a standard deviation of 0.01. For comparison, the performance of MLPs trained using He initialization was also measured.

In addition, Grad-CAM was used to visualize the reasoning behind the ENN's decisions. Heatmaps representing the decision basis of the ENNs were generated during the ENN B1's and B2's inference based on the gradients from the last convolutional layers.

$$lr_2 = lr_1 \frac{10^{-3}}{|\bar{y}|} \quad \ldots (4)$$

$$\Delta W = \begin{cases} -lr_1 \nabla L + lr_2 \nabla ENN(W) & if\ b_t \%10 = 0 \\ -lr_1 \nabla L & if\ b_t \%10 \neq 0 \end{cases} \quad \ldots (5)$$

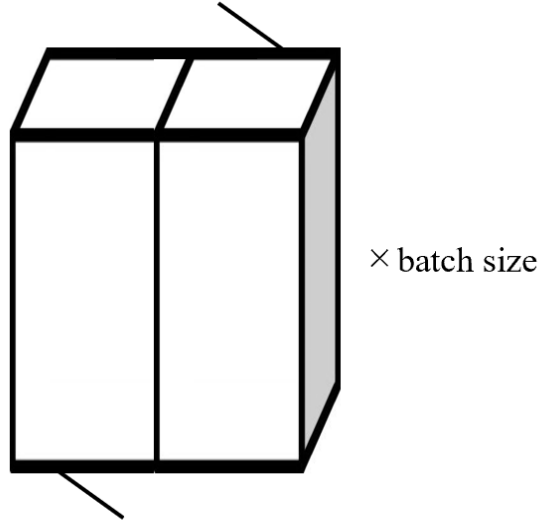

Fig. 3.1.1. Array data input to ENN

## 3.2. Analysis

In many cases using the proposed method, the test accuracy was approximately 2% higher than when using Vanilla MLP or L1 Regularization. Particularly when using ENN B1 with high-resolution input images, the test accuracy was about 2.8% higher than that of the Vanilla MLP, suggesting that the higher resolution of input data contributes to performance improvement. It was also found that sufficient predictions could be made even with reduced input image resolution.

Even though the test accuracy when using ENN B1 was only 0.28% higher compared to using He initialization, the difference between test error and training error was significantly smaller with ENN B1, indicating that ENN was able to optimize without inducing overfitting. However, the MLP trained using ENN B2 that with high resolution had a test accuracy of 90.606, which was almost the same as the Vanilla MLP's 90.211. This suggests that ENN B2 itself have been overfitting. The difference between test error and training error in ENN B2 was about 36 higher than B1's -6.089, clearly showing an overfitting tendency. From this result, it can be inferred that by recording the performance of ENNs during training and the types of data learned, optimal ENN for training of particular NN can be predicted.

Figure 3.2.1 illustrates the learning process when using ENN B1. The top three diagrams show the hidden layers at the beginning of training (epoch 1), and the bottom three diagrams show the hidden layers towards the end of training (epoch 24). From this figure, it can be seen that the patterns of gradients from the ENN in the early stages resemble the patterns of the weight parameters in the later stages. This suggests that ENN B1 played a significant role in forming the overall patterns of the weight parameters. Additionally, the heatmap at the end of training appears to partially match the pattern of the weight parameters in the same epoch. Particularly, a ring-shaped pattern appeared along the periphery of the region where large weight parameters were concentrated in the lower center of the heatmap at the end of training. This indicates that ENN B1 focused on the regions with large absolute values of the target MLP's parameters, judging that these regions contributed to the test accuracy of the target system.

Figure 3.2.2 shows the training process using ENN B1 at the top and ENN B2 at the bottom, highlighting the differences in behavior between ENN B1, which significantly improved the MLP, and ENN B2, which exhibited overfitting tendency. Towards the end of training (right side of the figure), the gradients from the ENN B1 during BP were generally smaller compared to those of ENN B2. This suggests that ENN B1 initially made significant improvements with large gradients but proceeded with more cautious, smaller gradients towards the end of training. Furthermore, according to the heatmap, ENN B1 evaluated a relatively narrow range highly towards the end of training, while ENN B2 evaluated a broader range highly. Therefore, it might be easier to predict and improve the performance of MLPs by considering that only a narrow region generally contributes to the accuracy of the MLP. According to the Strong Lottery Ticket Hypothesis (Jonathan Frankle & Michael Carbin, 2019), the weights of MLPs become divided into important and unimportant ones during training. The results of the heatmap from ENN B1 support this hypothesis.

Table 3.2.1 Generalizability of each method on the MNIST dataset (30 epochs of training)

| methods | MLP's test accuracy | MLP's test loss - train loss | ENN's perf | ENN's test loss - train loss | input size |
|---|---|---|---|---|---|
| Vanilla | 90.211 | 0.03 | - | - | - |
| He initialization | 92.736 | 2.183 | - | - | |
| L1 Regularization | 90.417 | 0.023 | - | - | |
| ENN S1 | 92.656 | **-0.067** | 98.302 | 2.354 | $\times 0.1^2$ |
| ENN S2 | 92.088 | 0.01 | 98.759 | -1.691 | |
| ENN B1 | **93.021** | 0.17 | 98.895 | **-6.089** | $\times 0.15^2$ |
| ENN B2 | 90.606 | 0.185 | 99.27 | 30.015 | |

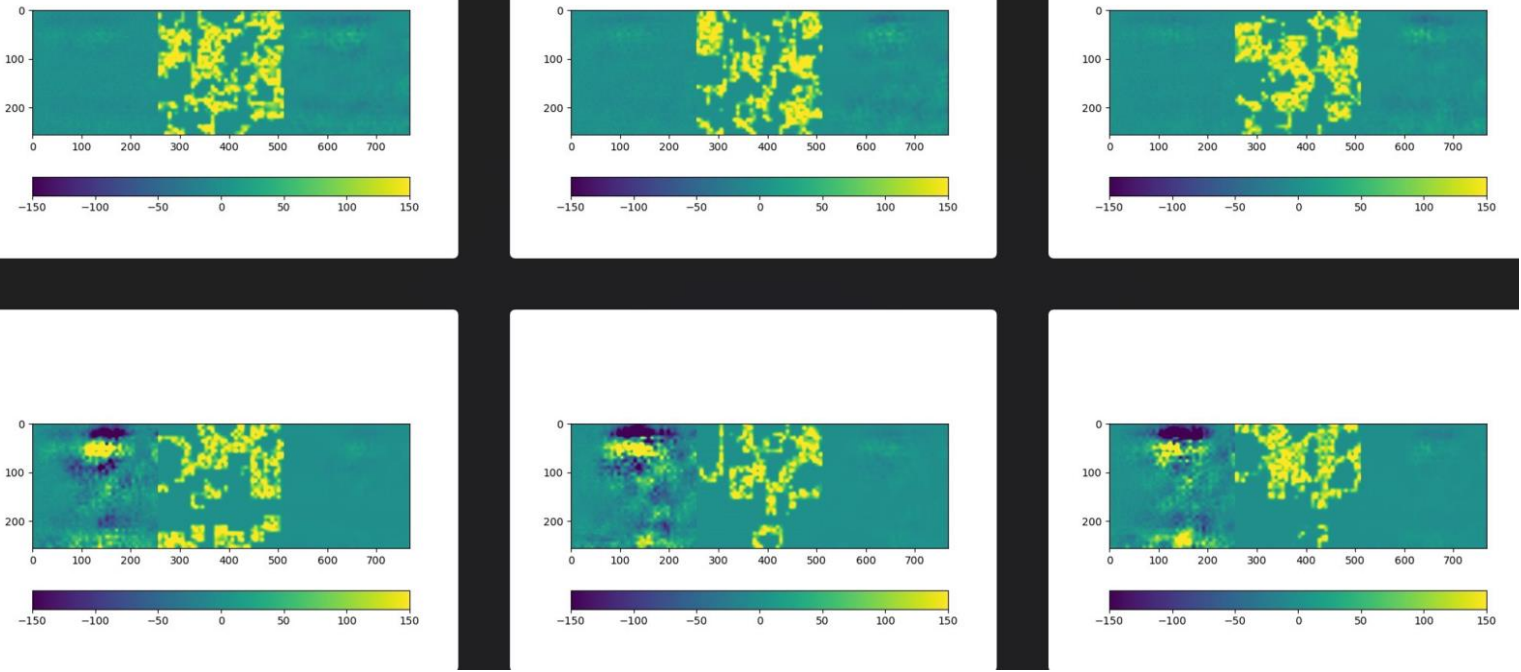

Fig. 3.2.1 MLP's 3 hidden layers at epoch 1 (top) and 24 (bottom) when using ENN B1

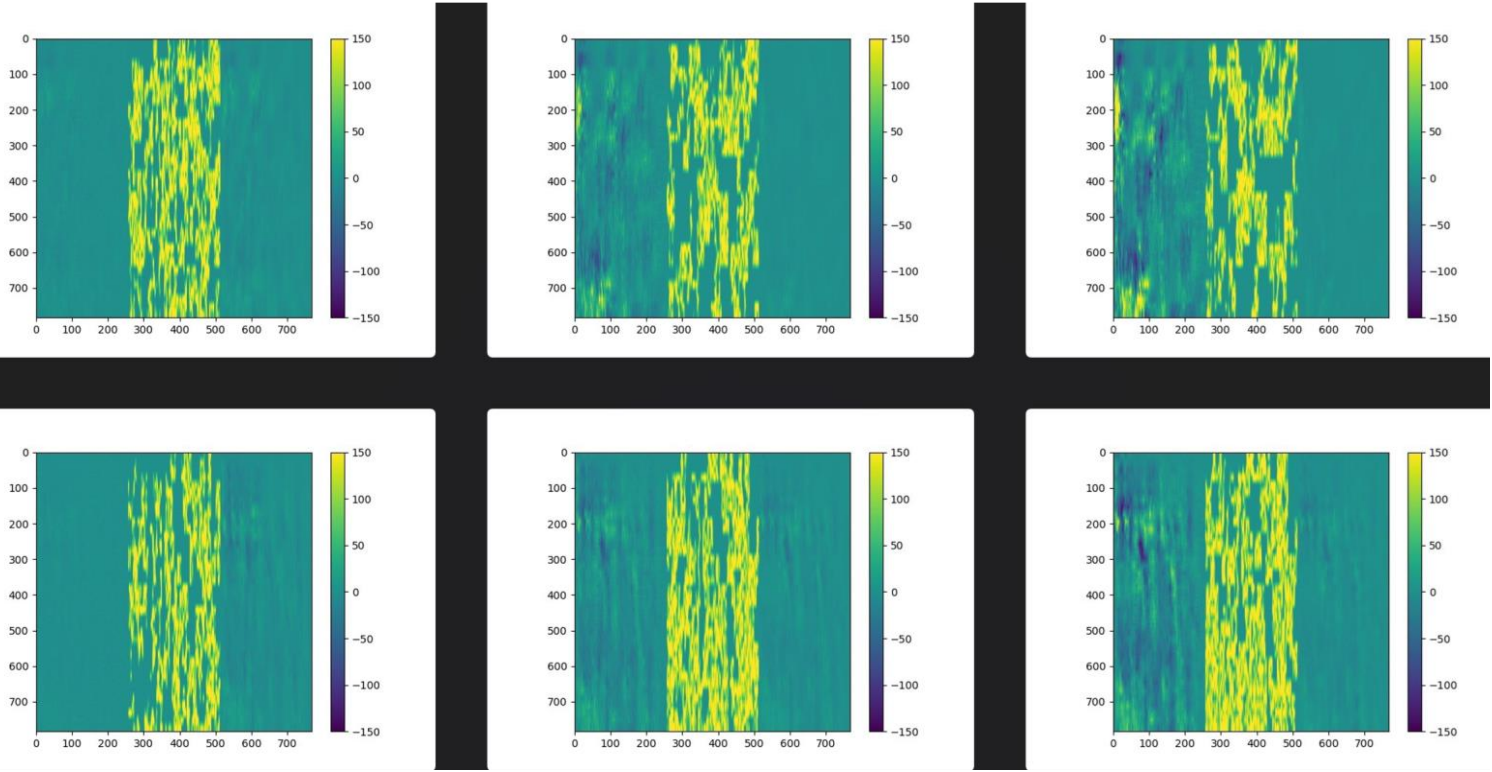

Fig. 3.2.2 MLP's input layers at epoch 1 (left), 12 (center), 24 (right) when using ENN B1 (top), B2 (bottom)

## 4. Conclusion

ENNs can learn to predict the performance of the target NN and apply this to rapidly explore optimal parameters. It was also found that MLPs trained with the proposed method improved accuracy without overfitting.

Additionally, ENNs can adjust the magnitude of gradients, and selecting an ENN that is not overfitted can significantly improve the target. Furthermore, it was found that the target can be improved even when the resolution of the data input to the ENN is reduced, and the evaluation of each parameter of the target by the ENN can be visualized.

## 5. future outlook on research

In this study, only data from each layer of the MLPs was used to train the ENNs, but I believe that by increasing the variety of NNs used as datasets, more general rules about NNs can be discovered. Furthermore, by miniaturizing ENNs through techniques such as Knowledge Distillation (Jianping Gou et al., 2021) and Pruning (Davis Blalock et al., 2020), it is possible to predict a greater number of phenomena with fewer computations. I believe that this implies that research activities traditionally performed by humans could be partially streamlined using ENNs.

For example, in physics research, there is a trend towards discovering equations that explain more phenomena and are easier to compute. These equations are based on empirically obtained knowledge, which parallels the empirical learning and inference application of ENNs, the miniaturization of NNs with current technology, and the visualization of decision bases.

In this study, the ENNs were used to improve MLPs, a type of NN. However, it seems likely that with appropriate training, ENNs could improve non-NN targets (e.g., the dimensions of components in robots). This is because ENNs can predict rewards in many situations given sufficient data about the target, and the amount of data required for training can be reduced by having another ENN improve the ENN itself. Transfer learning and fine-tuning could also be considered as methods to reduce the training data required for ENNs.